\documentclass[10pt,twocolumn,letterpaper]{article}

\usepackage{cvpr}
\usepackage{times}
\usepackage{epsfig}
\usepackage{graphicx}
\usepackage{amsmath}
\usepackage{amssymb}

\usepackage{multirow}
\usepackage{graphicx}
\usepackage{subfigure}
\usepackage[linesnumbered, ruled]{algorithm2e}
\usepackage{algorithmic}
\cvprfinalcopy

\def\cvprPaperID{7243}

\ifcvprfinal\fi

\begin{document}

%%%%%%%%%%%%%%%%%%%%%%%%%%%%%%%%%%%%%%%% Title %%%%%%%%%%%%%%%%%%%%%%%%%%%%%%%%%%%%%%%%
\title{Label Decoupling Framework for Salient Object Detection}
\author{
   Jun Wei\textsuperscript{\rm 1,2}, 
   Shuhui Wang\textsuperscript{\rm 1}\thanks{Corresponding author} ,
   Zhe Wu\textsuperscript{\rm 2,3}, 
   Chi Su\textsuperscript{\rm 4}, 
   Qingming Huang\textsuperscript{\rm 1,2,3}, 
   Qi Tian\textsuperscript{\rm 5}\\
   \textsuperscript{\rm 1}Key Lab of Intell. Info. Process., Inst. of Comput. Tech., CAS, Beijing, China\\
   \textsuperscript{\rm 2}University of Chinese Academy of Sciences, Beijing, China \quad
   \textsuperscript{\rm 3}Peng Cheng Laboratory, Shenzhen, China\\
   \textsuperscript{\rm 4}Kingsoft Cloud, Beijing, China  \quad
   \textsuperscript{\rm 5}Noah's Ark Lab, Huawei Technologies, China \\
   {\tt\small jun.wei@vipl.ict.ac.cn, wangshuhui@ict.ac.cn, zhe.wu@vipl.ict.ac.cn}\\
   {\tt\small suchi@kingsoft.com, qmhuang@ucas.ac.cn, tian.qi1@huawei.com}
}
\maketitle

%%%%%%%%%%%%%%%%%%%%%%%%%%%%%%%%%%%%%%% Abstract %%%%%%%%%%%%%%%%%%%%%%%%%%%%%%%%%%%%%%
\begin{abstract}
  To get more accurate saliency maps, recent methods mainly focus on aggregating multi-level features from fully convolutional network (FCN) and introducing edge information as auxiliary supervision. Though remarkable progress has been achieved, we observe that the closer the pixel is to the edge, the more difficult it is to be predicted, because edge pixels have a very imbalance distribution. To address this problem, we propose a label decoupling framework (LDF) which consists of a label decoupling (LD) procedure and a feature interaction network (FIN). LD explicitly decomposes the original saliency map into body map and detail map, where body map concentrates on center areas of objects and detail map focuses on regions around edges. Detail map works better because it involves much more pixels than traditional edge supervision. Different from saliency map, body map discards edge pixels and only pays attention to center areas. This successfully avoids the distraction from edge pixels during training. Therefore, we employ two branches in FIN to deal with body map and detail map respectively. Feature interaction (FI) is designed to fuse the two complementary branches to predict the saliency map, which is then used to refine the two branches again. This iterative refinement is helpful for learning better representations and more precise saliency maps. Comprehensive experiments on six benchmark datasets demonstrate that LDF outperforms state-of-the-art approaches on different evaluation metrics. Codes can be found at \url{https://github.com/weijun88/LDF}.
\end{abstract}

%%%%%%%%%%%%%%%%%%%%%%%%%%%%%%%%%%%% Introduction %%%%%%%%%%%%%%%%%%%%%%%%%%%%%%%%%%%%
\section{Introduction}
\label{introduction}

\begin{table}[htb]
  \caption{Mean absolute error of the predicted saliency maps (MAE$_{global}$) and edge areas (MAE$_{edge}$) of two state-of-the-art methods over three datasets. MAE$_{edge}$ is much larger than MAE$_{global}$, demonstrating that edge prediction is more difficult.}
  \label{Saliency&Edge}
  \small
  \renewcommand\tabcolsep{2.9pt}
  \renewcommand\arraystretch{1}
  \begin{tabular}{c|ccc|ccc}
    \hline
    \hline
    \multirow{2}{*}{} & \multicolumn{3}{c|}{EGNet~\cite{EGNet}} & \multicolumn{3}{c}{SCRN~\cite{SCRN}}\\ 
                            &  ECSSD & DUTS & DUT-O & ECSSD & DUTS & DUT-O \\
    \hline
    $MAE_{global}$  & 0.037 & 0.039 & 0.053 & 0.037 & 0.040 & 0.056\\
    $MAE_{edge}$  & 0.289 & 0.292 & 0.298 & 0.299 & 0.297 & 0.302\\
    \hline
    \hline
  \end{tabular}
\end{table}

%\vspace{-2cm} 

\begin{figure}[htb]
  \centering
  \subfigure[EGNet~\cite{EGNet}]{\includegraphics[width=0.48\linewidth]{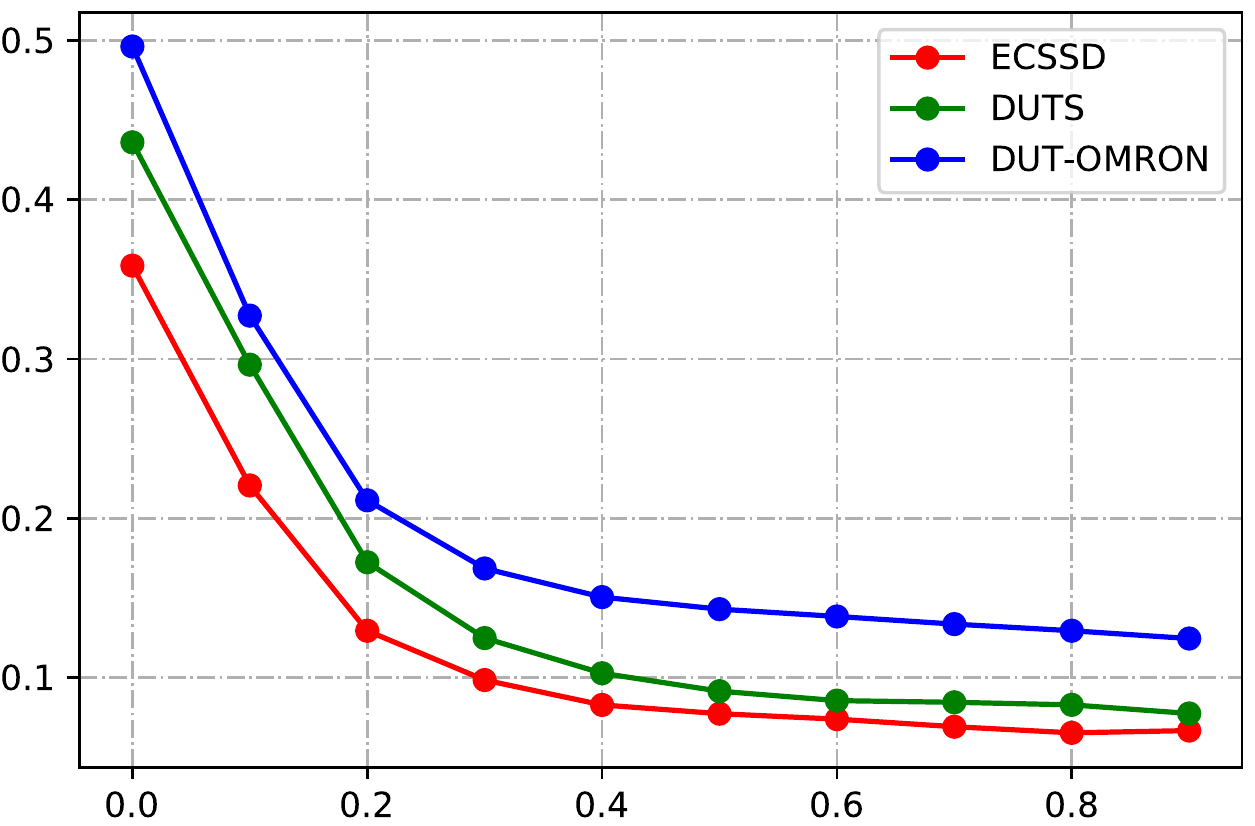}}
  \subfigure[ SCRN~\cite{SCRN} ]{\includegraphics[width=0.48\linewidth]{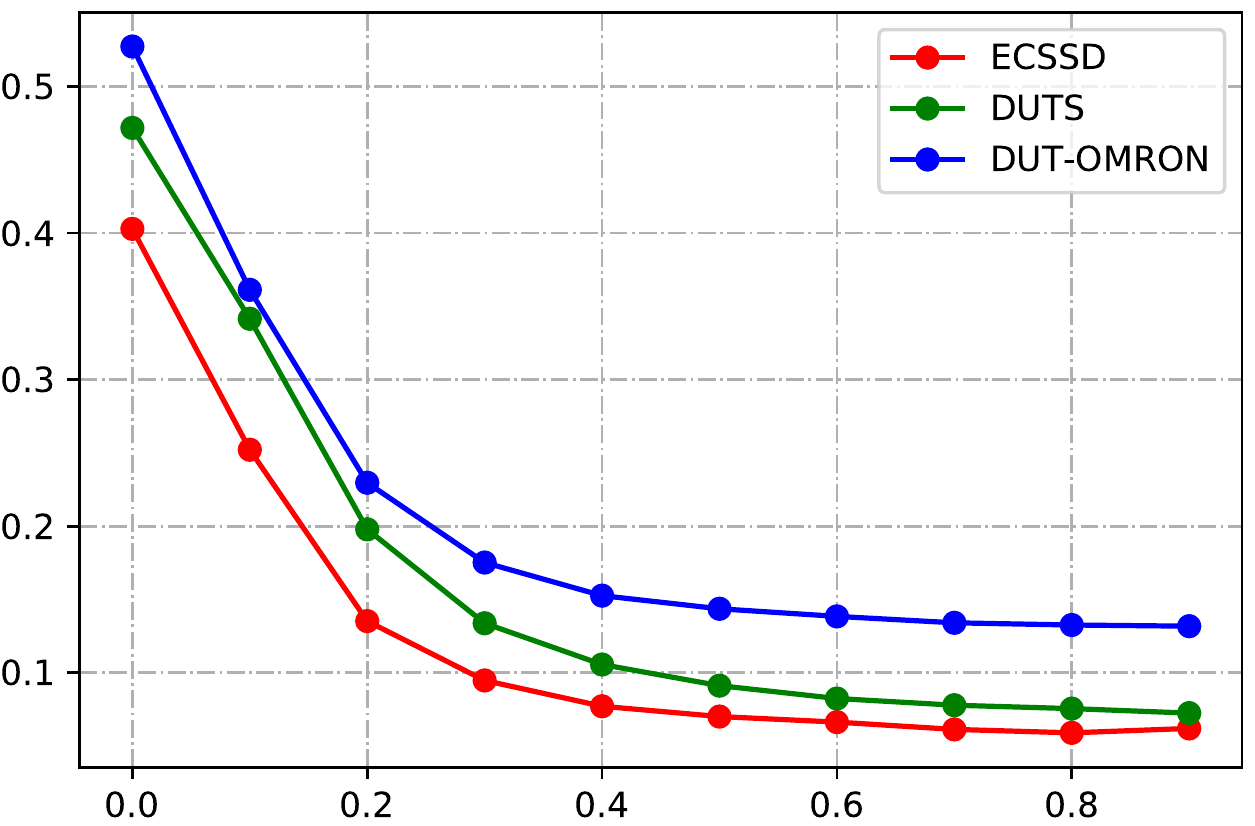}}
  \caption{Distribution of prediction error with respect to distance from pixel to its nearest edge. Horizontal coordinate represents the distance, which has been normalized to [0,1] and vertical coordinate is the prediction error. As can be seen, the closer the pixel is to the edge, the more difficult it is to be predicted.}
  \label{error&distance}
 \end{figure}
%%%%%%%%%%%%%%%%%%%%%%%%%%%%%%%%%%%%%%%%%%%%%%%%%%%%%%%%%%%%%%%%%%%%%%%%%%%%%%%%%%%%%%%%%%%%%%%%%%%%%%
Salient object detection (SOD)~\cite{AchantaHES09,ChengMHTH15,FanWCS19,Fan_2018_ECCV,Fan_2019_CVPR} aims at identifying the most visually attractive objects or parts in an image or video, which is widely applied as a pre-processing procedure in downstream computer vision tasks~\cite{Survey, xin2018reverse}. During the past decades, researchers have proposed hundreds of SOD methods based on hand-crafted features ({\it e.g.,} color, texture and brightness)~\cite{Survey}. However, these features can not capture high-level semantic information, which restricts their applications in complex scenes. Recently, convolutional neural networks (CNNs) have demonstrated powerful capability of feature representation and greatly promoted the development of SOD. Many CNNs-based methods~\cite{DSS, Amulet, RFCN,SRM, DGRL, PAGR, RAS, PICANet, R3Net, PFA, RADF, MLMSNet} have achieved remarkable performance by designing different decoders to aggregate multi-level CNN features. To get better feature representations, these methods focus on mining more context information and devising more effective feature fusion strategies. Besides, introducing the boundary information is another key point in SOD. Existing methods attempt to take edges as supervision to train SOD models, which significantly improves the accuracy of saliency maps \cite{BASNet,PoolNet,EGNet,SCRN,SIBA,AFNet}. 

However, the imbalance between edge pixels and non-edge ones makes it hard to get good edge predictions. Therefore, directly taking edges as supervision may lead to suboptimal solutions. To better elaborate this statement, we calculate the mean absolute error (MAE) of two state-of-the-art methods ({\it i.e.,} EGNet~\cite{EGNet} and SCRN~\cite{SCRN}) over three SOD datasets ({\it i.e.,} ECSSD~\cite{ECSSD}, DUTS~\cite{DUTS} and DUT-O~\cite{DUTO}) in Tab.~\ref{Saliency&Edge}. Though two methods get low error in global saliency prediction, they perform much worse in edge prediction, which shows that edge pixels are more difficult to predict than others. To further explore the prediction difficulties of pixels, we analyse the distribution of prediction error about the distance to the nearest edge of EGNet and SCRN in Fig.~\ref{error&distance}. 

In Fig.~\ref{error&distance}, the prediction error curves gradually increases from far away to close to the edge ({\it i.e.,} the right axis to the left axis). When the distance is larger than 0.4, these curves rise slowly. However, when the distance gets smaller than 0.4, these curves begin to go upwards quickly. Based on this observation, we can divide each of the curves into two parts according to pixel distance from their nearest edges. Pixels near the edges correspond to much larger prediction errors than far-away pixels. These pixels with high prediction errors consists of both edge pixels and many other pixels close to edges that are ignored by recent edge-aware methods. Most of the hard pixels that can greatly improve the performance of SOD are not fully used, while using only edge pixels will lead to difficulties because of the imbalance distribution between edge pixels and background ones. In contrast, pixels far away from edges have relatively low prediction errors, which are much easier to be classified. However, traditional saliency labels treat all pixels inside salient object equally, which may cause pixels with low prediction errors to suffer distractive effects from those near edges. 

We propose label decoupling framework to address the above problems. LDF mainly consists of a label decoupling procedure and a feature interaction network. As shown in Fig.~\ref{sketch}, a saliency label is decomposed into a body map and a detail map by LD. Different from the pure edge map, the detail map consists of both edges as well as nearby pixels, which makes full use of pixels near edge and thus has a more balanced pixel distribution. 
The body map mainly concentrates on pixels far away from edges. Without the disturbance of pixels near edges, the body map can supervise the model to learn better representations. Accordingly, FIN is designed with two branches to adapt to body map and detail map respectively. 
The two complementary branches in FIN are fused to predict the saliency map, which is then used to refine the two branches again. This iterative refinement procedure is helpful for obtaining gradually accurate saliency maps prediction.

We conduct experiments on six popular SOD datasets and demonstrate the superior performance of LDF. In summary, our contributions are as follows: 
\begin{itemize}
  \item We analyse the shortcomings of edge-based SOD methods and propose a label decoupling procedure to decompose a saliency label into body map and detail map to supervise the model, respectively. 
  \item We design a feature interaction network to make full use of the complementary information between branches. Both branches will be enhanced by iteratively exchanging information to produce more precise saliency maps.
  \item Extensive experiments on six SOD datasets show that our model outperforms state-of-the-art models by a large margin. In particularly, we demonstrate the good performance of LDF in different challenging scenes in the SOC dataset~\cite{SOC}.
\end{itemize}

\section{Related Work}
\begin{figure*}[htb]
  \centering
  \includegraphics[scale=0.515]{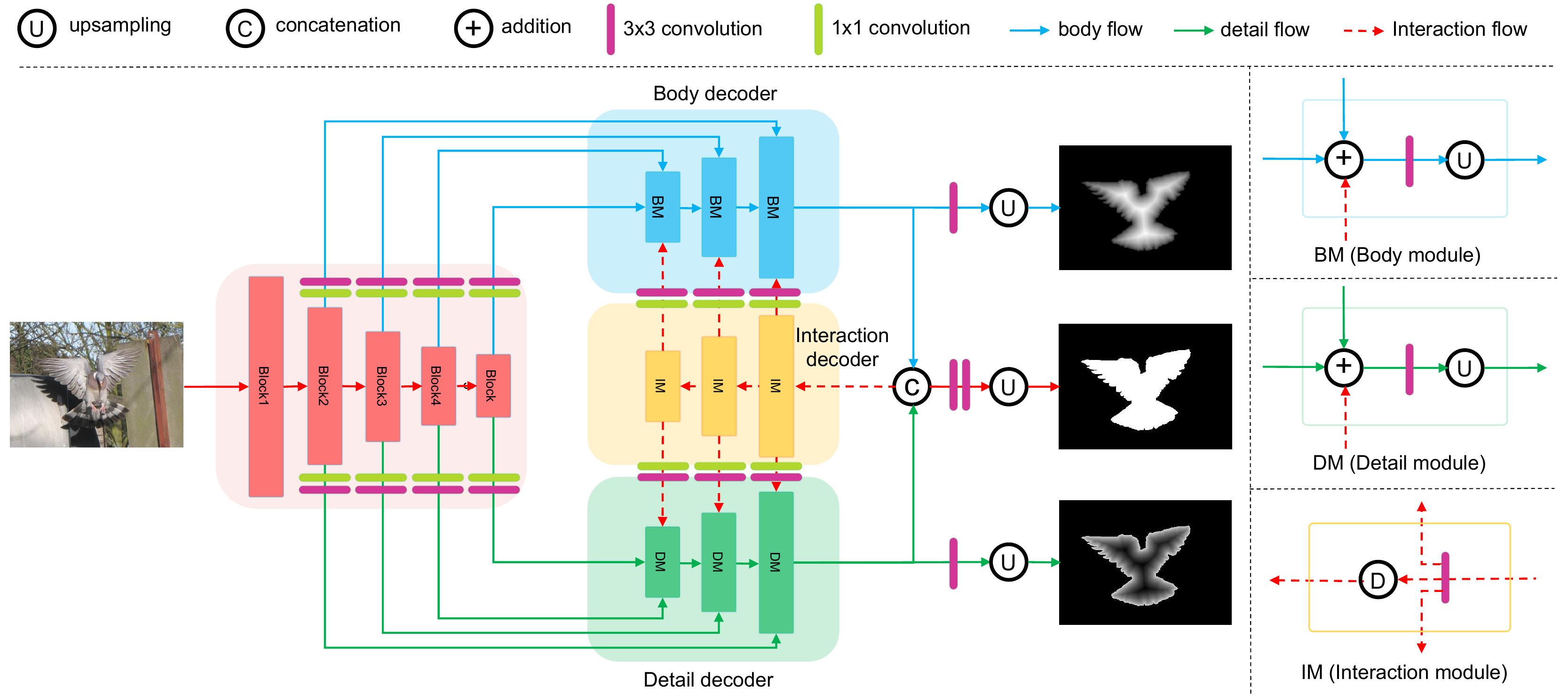}
  \caption{An overview of our proposed label decoupling framework (LDF). LDF is based on ResNet-50~\cite{Resnet} with supervision from body map, detail map and saliency map. LDF consists of two encoders and two decoders, {\it i.e.}, a backbone encoder for feature extraction, an interaction encoder for exchanging information, a body decoder and a detail decoder to generate body map and detail map respectively. The interaction encoder is not involved until body decoder and detail decoder output features.}
  \label{framework}
 \end{figure*}
%%%%%%%%%%%%%%%%%%%%%%%%%%%%%%%%%%%%%%%%%%%%%%%%%%%%%%%%%%%%%%%%%%%%%%%%%%%%%%%%%%%%%%%%%%%%%%%%%%%%%%
During the past decades, a huge body of traditional methods have been developed for SOD. These methods~\cite{WangJYCHZ17, BorjiI12, ECSSD} mainly rely on intrinsic cues ({\it e.g.}, color and texture) to extract features. However, these features cannot capture high-level semantic information and are not robust to variations, which limits their applications in complex scenarios. Recently, deep learning based models have achieved remarkable performance, which can be divided into aggregation-based models and edge-based models.

\subsection{Aggregation-based Models}
Most of the aggregation-based models adopt the encoder-decoder framework, where the encoder is used to extract multi-scale features and the decoder is used to integrate the features to leverage context information of different levels. Hou {\it et al.}~\cite{DSS} constructed shortcut connections on fully convolutional networks~\cite{FCN} and integrated features of different layers to output more accurate maps. Chen {\it et al.}~\cite{RAS} proposed a reverse attention network, which erased the current predicted salient regions to expect the network to mine out the missing parts. Deng {\it et al.}~\cite{R3Net} designed an iterative strategy to learn the residual map between the prediction and ground truth by combining features from both deep and shallow layers. Wu {\it et al.}~\cite{CPD} found that features of shallow layers greatly increased the computation cost, but only brought  little improvement in final results. Liu {\it et al.}~\cite{PoolNet} utilized simple pooling and a feature aggregation module to build fast and accurate model. Zhao {\it et al.}~\cite{PFA} introduced the channel-wise attention and spatial attention to extract valuable features and suppress background noise. Wang {\it et al.}~\cite{TDBU} designed a top-down and bottom-up workflow to infer the salient object regions with multiple iterations. Liu {\it et al.}~\cite{PICANet} proposed a pixel-wise contextual attention network to learn the context of each pixel, and combined the global context and local context for saliency prediction. Zhang {\it et al.}~\cite{BMPM} designed a bi-directional message passing model for better feature selection and integration.

\subsection{Edge-based Models}
In addition to saliency masks, edge label is also introduced to SOD in~\cite{BASNet, SCRN, PAGE, PoolNet, Amulet, PFA} to assist the generation of saliency maps. Zhang {\it et al.}~\cite{Amulet} and Zhao {\it et al.}~\cite{PFA} directly built the edge loss with binary cross-entropy to emphasize the importance of boundaries. Qin {\it et al.}~\cite{BASNet} designed a hybrid loss to supervise the training process of SOD on pixel-level, patch-level and map-level. Liu {\it et al.}~\cite{PoolNet} used additional edge dataset for joint training of both edge detection and SOD models. Feng {\it et al.}~\cite{AFNet} applied a boundary-enhanced loss to generate sharp boundaries and distinguish the narrow background margins between two foreground areas. Li {\it et al.}~\cite{C2SNet} used a two-branch network to simultaneously predict the contours and saliency maps, which can automatically convert the trained contour detection model to SOD model. Wu {\it et al.}~\cite{SCRN} investigated the logical inter-relations between segmentation and edge maps, which are then promoted to bidirectionally refine multi-level features of the two tasks. Although these methods take into account the relationship between edges and saliency maps, edge prediction is a hard task because of imbalanced pixel distribution. In this paper, we explicitly decouple the saliency label into body map and detail map, as shown in Fig.~\ref{sketch}. Detail map helps model learn better edge features and body map decreases the distraction from pixels near edges to center ones.

\section{Methodology}
In this section, we first introduce the label decoupling method and give the specific steps to decompose the saliency map into body map and detail map. Then, to take advantage of the complementarity between features, we introduce FIN which facilitates the iterative information exchange between branches. The overview of the proposed model is shown in Fig.~\ref{framework}.

%%%%%%%%%%%%%%%%%%%%%%%%%%%%%%%%%%%%%%%%%%%%%%%%%%%%%%%%%%%%%%%%%%%%%%%%%%%%%%%%%%%%%%%%%%%%%%%%%%%%%%
\subsection{Label Decoupling}
\begin{figure}
  \centering
  \includegraphics[scale=0.57]{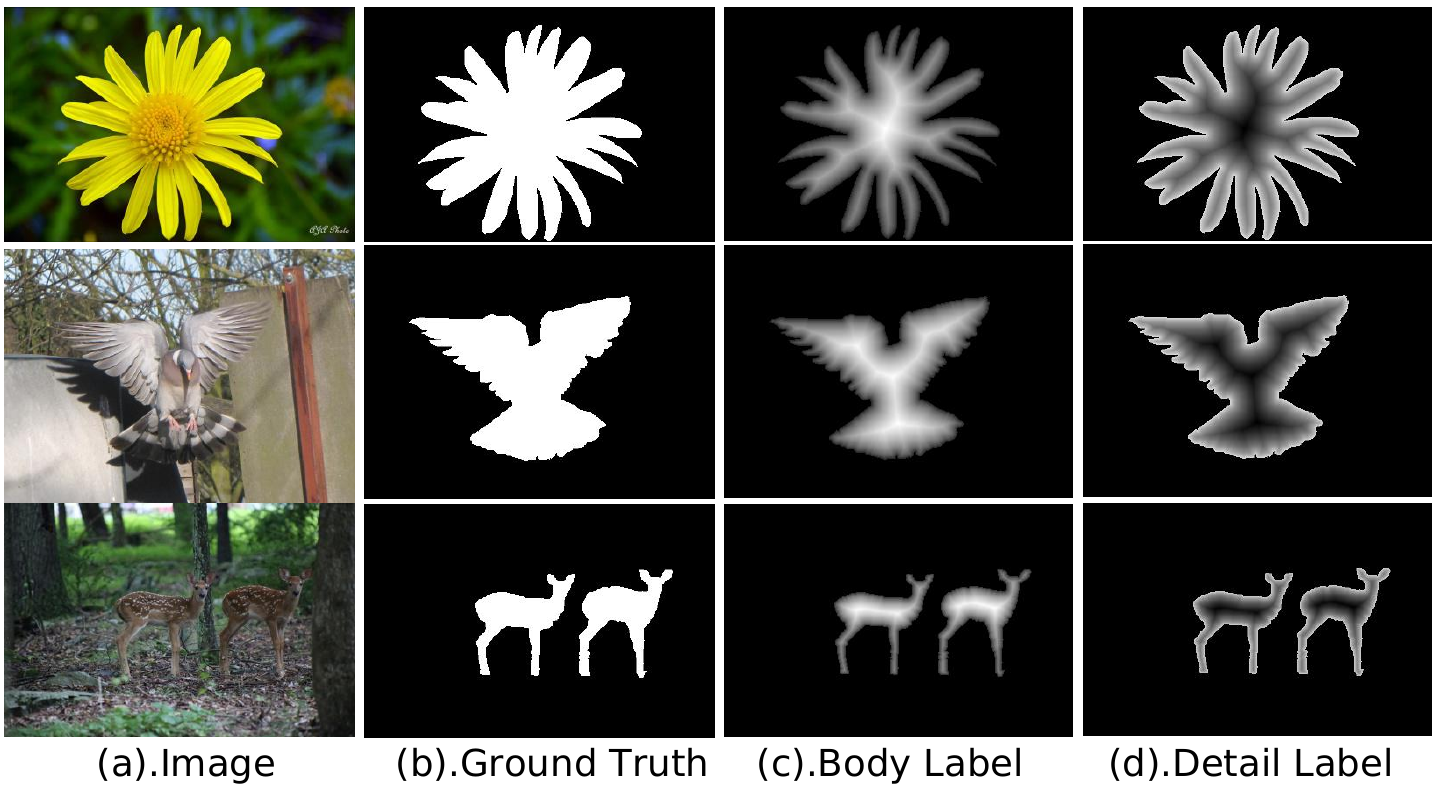}
  \caption{Some examples of label decoupling. (c) represents the body label of the ground truth, where pixels close to the center of the target have larger values. (d) means the detail label of the ground truth, where pixels near the boundary of the target have larger values. The sum of (c) and (d) is equal to (b).}
  \label{sketch}
 \end{figure}

As described in Sec.~\ref{introduction}, the prediction difficulty of a pixel is closely related to its position. Because of the cluttered background, pixels near the edge are more prone to be mispredicted. In comparison, central pixels have higher prediction accuracy due to the internal consistency of the salient target. Instead of treating these pixels equally, it will be more reasonable to deal with them according to their respective characteristics. Accordingly, we propose to decouple the original label into body label and detail label, as shown in Fig.~\ref{sketch}. To achieve this goal, we introduce Distance Transformation (DT) to decouple the original label, which is a traditional image processing algorithm. DT can convert the binary image into a new image where each foreground pixel has a value corresponding to the minimum distance from the background by a distance function. 

Specifically, the input of DT is a binary image $I$, which can be divided into two groups ({\it i.e.}, foreground $I_{fg}$ and background $I_{bg}$. For each pixel $p$, $I(p)$ is its corresponding value. If $p \in I_{fg}$, $I(p)$ equals 1, and 0 if $p \in I_{bg}$. To get the DT result of image $I$, we define the metric function $f(p, q)=\sqrt{(p_x-q_x)^2+(p_y-q_y)^2}$ to measure the distance between pixels. If pixel $p$ belongs to the foreground, DT will first look up its nearest pixel $q$ in the background and then use $f(p,q)$ to calculate the distance between pixel $p$ and $q$. If pixel $p$ belongs to the background, their minimum distance is set to zero. We use $f(p,q)$ as the pixels of a newly generated image, and the distance transformation can be expressed as

\begin{flalign}
  I^{'}(p) &= \left\{
                \begin{aligned}
                  \min\limits_{q \in I_{bg}} f(p, q), \qquad p \in I_{fg} \\
                  0, \qquad p \in I_{bg}
                \end{aligned}
              \right. \label{DistTrans}
\end{flalign}

After the distance transformation, the original image $I$ has been transformed into $I^{'}$ where pixel value $I^{'}(p)$ no longer equals to 0 or 1. We normalize the pixel values in $I^{'}$ using a simple linear function $I^{'}=\frac{I^{'}-min(I^{'})}{max(I^{'})-min(I^{'})}$ to map the original value to [0, 1]. Compared with the original image $I$ which treats all pixels equally, pixel value of $I^{'}$ not only depends on whether it belongs to foreground or background, but also is related to its relative position. Pixels located in the center of object have the largest values and those far away from the center or in background have the smallest values. So $I^{'}$ represents the body part of the original image, which mainly focuses on the central pixels that are relatively easy. We use it as the body label in the following experiments. Correspondingly, by removing the body image $I^{'}$ from the original image $I$, we can get the detail image, which is regarded as the detail label in consequent experiments and mainly concentrates on pixels far away from the main regions. In addition, we multiply the newly generated labels with the original binary image $I$ to remove the background interference as

\begin{flalign}
  Label\Rightarrow \left\{
                  \begin{aligned}
                    BL &= I*I^{'} \\
                    DL &= I*(1-I^{'}) 
                  \end{aligned} 
                \right. \label{detail}
\end{flalign}
where $BL$ means the body label and $DL$ represents the detail label. Now the original label has been decoupled into two different kinds of supervision to assist the network to learn both the body and detail features with different characteristics respectively.

\subsection{Feature Extraction}
\label{featureextraction}

As suggested by~\cite{DGRL,SRM,PICANet}, we use ResNet-50~\cite{Resnet} as our backbone network. Specifically, we remove the fully connected layer and retain all convolutional blocks. Given an input image with shape $H \times W$, this backbone will generate five scales of features with decreasing spatial resolution by stride 2 due to downsampling. We denote these features as $F=\{F_i | i=1,2,3,4, 5\}$. The size of the $i$-th feature is $\frac{W}{2^i}\times \frac{H}{2^i} \times C_i$, where $C_i$ is the channel of the $i$-th feature. It has been shown that low-level features greatly increase computation cost, but bring limited performance improvement~\cite{CPD}. So we only utilize features from $\{F_i | i=2,3,4,5\}$, as shown in Fig.~\ref{framework}. Two convolution layers are applied to these features to adapt them seperately to the body prediction task and detail prediction task.  Then we get two groups of features $B=\{B_i | i=2,3,4,5\}$ and $D=\{D_i | i=2,3,4,5\}$, which all have been squeezed to 64 channels and sent to the decoder network for saliency map generation.

\subsection{Feature Interaction Network}
Feature interaction network is built to adapt to the label decoupling, as shown in Fig.~\ref{framework}. With label decoupling, the saliency label has been transformed into the body map and the detail map, both of which are taken as supervision for model learning. FIN is designed as a two-branch structure, each of which is responsible for one label kind. Since both the body map and detail map are derived from the same saliency label, there exists a certain level of similarity and complementarity between the features from two branches. We introduce feature interaction between the complementary branches for information exchanging. 

On the whole, the proposed framework is made up of one backbone encoder network, one interaction encoder network, one body decoder network and one detail decoder network. As discussed in Sec.~\ref{featureextraction}, ResNet-50~\cite{Resnet} is used as the backbone network to extract multi-level features $B=\{B_i | i=2,3,4,5\}$ and $D=\{D_i | i=2,3,4,5\}$. For features $B$, a body decoder network is applied to generate body maps. Similarly, for features $D$, a detail decoder network is applied to generate detail maps. After getting the output features of these two branches, the simplest way to deal with them is to concatenate these features and apply a convolutional layer to get final saliency maps. However, this way ignores the relationship between branches. To explicitly promote the information exchange between branches, an interaction encoder network is introduced.

More specifically, interaction decoder takes the concatenated features of the body decoder and detail decoder as input. It stacks multiple convolutions to extract multi-level features. Then these multi-level features will be applied with 3x3 convolution layers to make them appropriate for body decoder and detail decoder respectively. Direct addition is used to fuse the interaction features with features from backbone encoder to produce more accurate saliency maps. On the surface, the whole network is unusual since the latter branch outputs are used in the former decoder. But in fact, feature interaction consists of multiple iterations. At the first iteration, two branches output features without exchanging information. From the second iteration, interaction is involved between branches.

\subsection{Loss Function} 
Our training loss is defined as the summation of the outputs of all iterations as,
\begin{small}
\begin{flalign}
    \label{totalloss}
    \mathcal{L} = \sum_{k=1}^K\alpha_k\ell^{(k)},
\end{flalign}
\end{small}
where $\ell^{(k)}$ is the loss of the $k$-th iteration, $K$ denotes the total number of iterations and $\alpha_k$ is the weight of each iteration. To simplify the problem, we set $\alpha_k=1$ to treat all iterations equally. For each iteration, we will get three outputs ({\it i.e.}, body, detail and segmentation) and each of them corresponds to one loss. So $\ell^{(k)}$ can be defined as the combination of three losses as follows:
\begin{flalign}
  \label{singleloss}
  \ell^{(k)} = \ell^{(k)}_{body}+\ell^{(k)}_{detail}+\ell^{(k)}_{segm},
\end{flalign}
where $\ell^{(k)}_{body}$, $\ell^{(k)}_{detail}$ and $\ell^{(k)}_{segm}$ denote body loss, detail loss and segmentation loss, respectively. We directly utilize binary cross entropy (BCE) to calculate both $\ell^{(k)}_{body}$ and $\ell^{(k)}_{detail}$. BCE is a widely used loss in binary classification and segmentation, which is defined as:
\begin{footnotesize}
  \begin{flalign}
    \label{bce}
    \ell_{bce}\!=\!-\!\sum\limits_{(x,y)}[g(x,\!y)log(p(x,\!y))\!+\!(1\!-\!g(x,\!y))log(1\!-\!p(x,\!y))],
  \end{flalign}
\end{footnotesize}
where $g(x,y) \in [0, 1]$ is the ground truth label of the pixel $(x,y)$ and $p(x,y) \in [0,1]$ is the predicted probability of being salient object. However, BCE calculates the loss for each pixel independently and ignores the global structure of the image. To remedy this problem, as suggested by~\cite{BASNet} we utilize the IoU loss to calculate $\ell^{(k)}_{segmentation}$, which can measure the similarity of two images on the whole rather than a single pixel. It is defined as:
\begin{small}
\begin{eqnarray}
  \label{iou}
  \ell_{iou} = 1-\frac{\sum\limits_{(x,y)} [g(x,y)*p(x,y)]}{\sum\limits_{(x,y)} [g(x,y)+p(x,y)-g(x,y)*p(x,y)]},
\end{eqnarray}
\end{small}
where the notations are the same as Eq.~\ref{bce}. We do not apply IoU loss on $\ell^{(k)}_{body}$ and $\ell^{(k)}_{detail}$, because IoU loss requires the ground truth to be binary or it will result in wrong predictions, while body label and detail label do not satisfy this requirement.

\section{Experiments}
\begin{table*}
  \caption{Performance comparison with state-of-the-art methods on six datasets. MAE (smaller is better), mean $F$-measure ($mF$, larger is better) and $E$-measure ($E_\xi$, larger is better) are used to measure the model performance. '-' means the author has not provided corresponding saliency maps. The best and the second best results are highlighted in {\color{red}red} and {\color{blue}blue} respectively.}
  \label{Performance}
  \renewcommand\tabcolsep{2.35pt}
  \renewcommand\arraystretch{1}
  \centering
  \begin{tabular}{l|ccc|ccc|ccc|ccc|ccc|ccc}
     \hline
     \hline
     \multirow{3}{*}{\textbf{Algorithm}}  & \multicolumn{3}{c|}{\textbf{ECSSD}} & \multicolumn{3}{c|}{\textbf{PASCAL-S}} & \multicolumn{3}{c|}{\textbf{DUTS-TE}} & \multicolumn{3}{c|}{\textbf{HKU-IS}} & \multicolumn{3}{c|}{\textbf{DUT-OMRON}}  & \multicolumn{3}{c}{\textbf{THUR15K}}\\
      & \multicolumn{3}{c|}{1,000 images} & \multicolumn{3}{c|}{850 images} & \multicolumn{3}{c|}{5,019 images} & \multicolumn{3}{c|}{4,447 images} & \multicolumn{3}{c|}{5,168 images} & \multicolumn{3}{c}{6,232 images}\\
    \cline{2-19}
                              & MAE & $mF$ &$E_\xi$& MAE & $mF$ &$E_\xi$& MAE & $mF$ &$E_\xi$& MAE & $mF$ &$E_\xi$& MAE & $mF$ &$E_\xi$& MAE & $mF$ &$E_\xi$ \\
     \hline
     \hline
     BMPM~\cite{BMPM}         & .044 & .894 & .914 & .073 & .803 & .838 & .049 & .762 & .859 & .039 & .875 & .937 & .063 & .698 & .839 & .079 & .704 & .803 \\
     DGRL~\cite{DGRL}         & .043 & .903 & .917 & .074 & .807 & .836 & .051 & .764 & .863 & .037 & .881 & .941 & .063 & .709 & .843 & .077 & .716 & .811 \\
     R$^3$Net~\cite{R3Net}    & .051 & .883 & .914 & .101 & .775 & .824 & .067 & .716 & .827 & .047 & .853 & .921 & .073 & .690 & .814 & .078 & .693 & .803 \\
     RAS~\cite{RAS}           & .055 & .890 & .916 & .102 & .782 & .832 & .060 & .750 & .861 & .045 & .874 & .931 & .063 & .711 & .843 & .075 & .707 & .821 \\
     PiCA-R~\cite{PICANet}    & .046 & .867 & .913 & .075 & .776 & .833 & .051 & .754 & .862 & .043 & .840 & .936 & .065 & .695 & .841 & .081 & .690 & .803 \\
     AFNet~\cite{AFNet}       & .042 & .908 & .918 & .070 & .821 & .846 & .046 & .792 & .879 & .036 & .888 & .942 & .057 & .738 & .853 & .072 & .730 & .820 \\
     BASNet~\cite{BASNet}     & .037 & .880 & .921 & .076 & .775 & .847 & .048 & .791 & .884 & {\color{blue}.032} & .895 & .946 & .056 & {\color{blue}.756} & {\color{blue}.869} & .073 & .733 & .821 \\
     CPD-R~\cite{CPD}         & .037 & .917 & .925 & .072 & .824 & .849 & .043 & .805 & .886 & .034 & .891 & .944 & .056 & .747 & .866 & .068 & .738 & .829 \\
     EGNet-R~\cite{EGNet}     & .037 & .920 & .927 & .074 & .823 & .849 & {\color{blue}.039} & {\color{blue}.815} & .891 & {\color{blue}.032} & .898 & .948 & {\color{blue}.053} & .755 & .867 & .067 & {\color{blue}.741} & .829 \\
     PAGE~\cite{PAGE}         & .042 & .906 & .920 & .077 & .810 & .841 & .052 & .777 & .869 & .037 & .882 & .940 & .062 & .736 & .853 & - & - & - \\
     TDBU~\cite{TDBU}         & .041 & .880 & .922 & .071 & .779 & .852 & .048 & .767 & .879 & .038 & .878 & .942 & .061 & .739 & .854 & - & - & - \\
     SCRN~\cite{SCRN}         & .037 & .918 & .926 & {\color{blue}.064} & {\color{blue}.832} & {\color{blue}.857} & .040 & .808 & .888 & .034 & .896 & .949 & .056 & .746 & .863 & {\color{blue}.066} & {\color{blue}.741} & {\color{blue}.833} \\
     SIBA~\cite{SIBA}         & {\color{blue}.035} & {\color{blue}.923} & {\color{red}.928} & .070 & .830 & .855 & .040 & {\color{blue}.815} & {\color{blue}.892} & {\color{blue}.032 }& {\color{blue}.900} & {\color{blue}.950} & .059 & .746 & .860 & .068 & {\color{blue}.741} & .832 \\
     PoolNet~\cite{PoolNet}   & .039 & .915 & .924 & .074 & .822 & .850 & .040 & .809 & .889 & {\color{blue}.032} & .899 & .949 & .056 & .747 & .863 & .070 & .732 & .822 \\
     \hline
     \textbf{LDF(ours)}       & {\color{red}.034} & {\color{red}.930} & {\color{blue}.925} & {\color{red}.060} & {\color{red}.848} & {\color{red}.865} & {\color{red}.034} & {\color{red}.855} & {\color{red}.910} & {\color{red}.027} & {\color{red}.914} & {\color{red}.954} & {\color{red}.051} & {\color{red}.773} & {\color{red}.873} & {\color{red}.064} & {\color{red}.764} & {\color{red}.842} \\
     \hline
     \hline
  \end{tabular}
\end{table*}

\begin{figure*}
 \centering
 \includegraphics[scale=0.505]{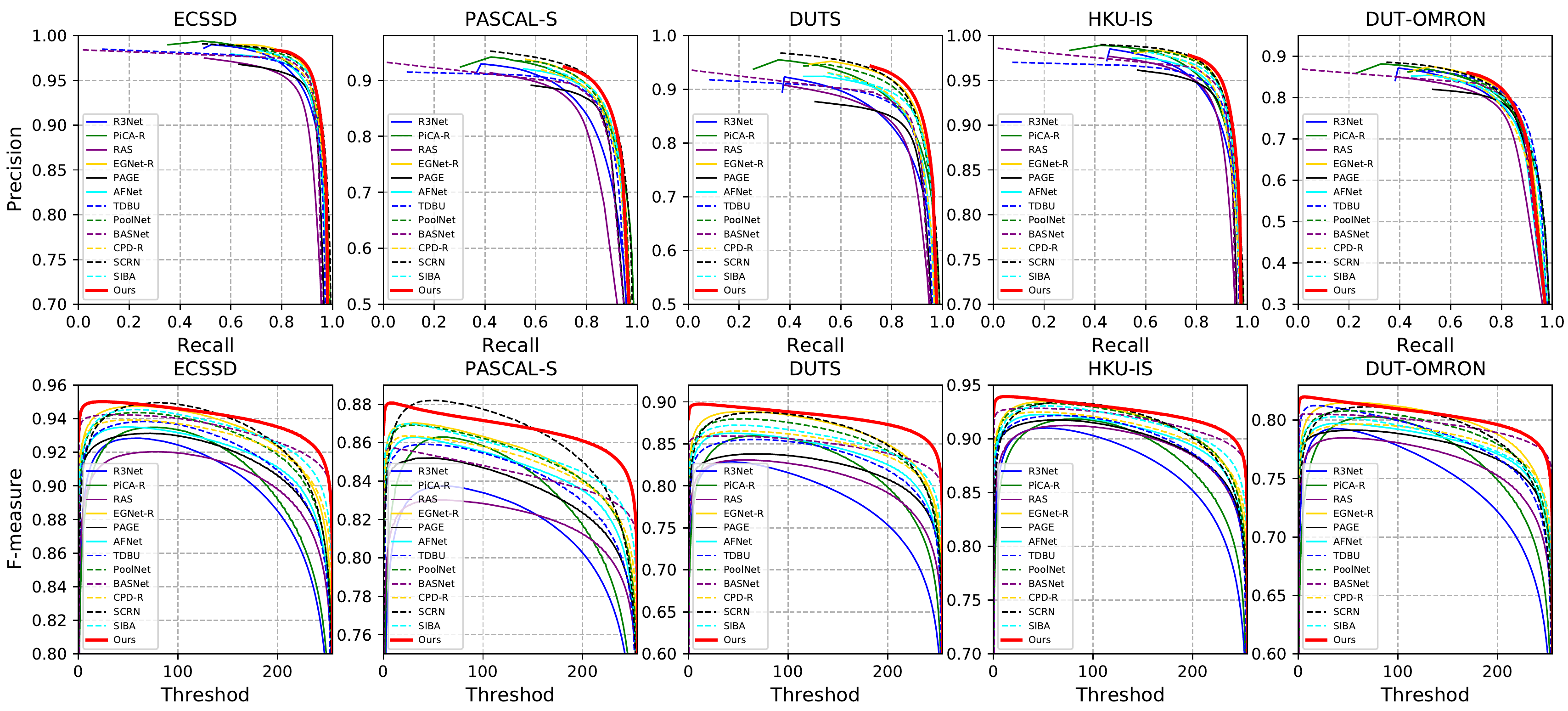}
 \caption{Performance comparison with state-of-the-art methods on five datasets. The first row shows precision-recall curves. The second row shows $F$-measure curves with different thresholds.}
 \label{PRCurve}
\end{figure*}

\begin{figure*}[htb]
  \centering
  \includegraphics[scale=0.515]{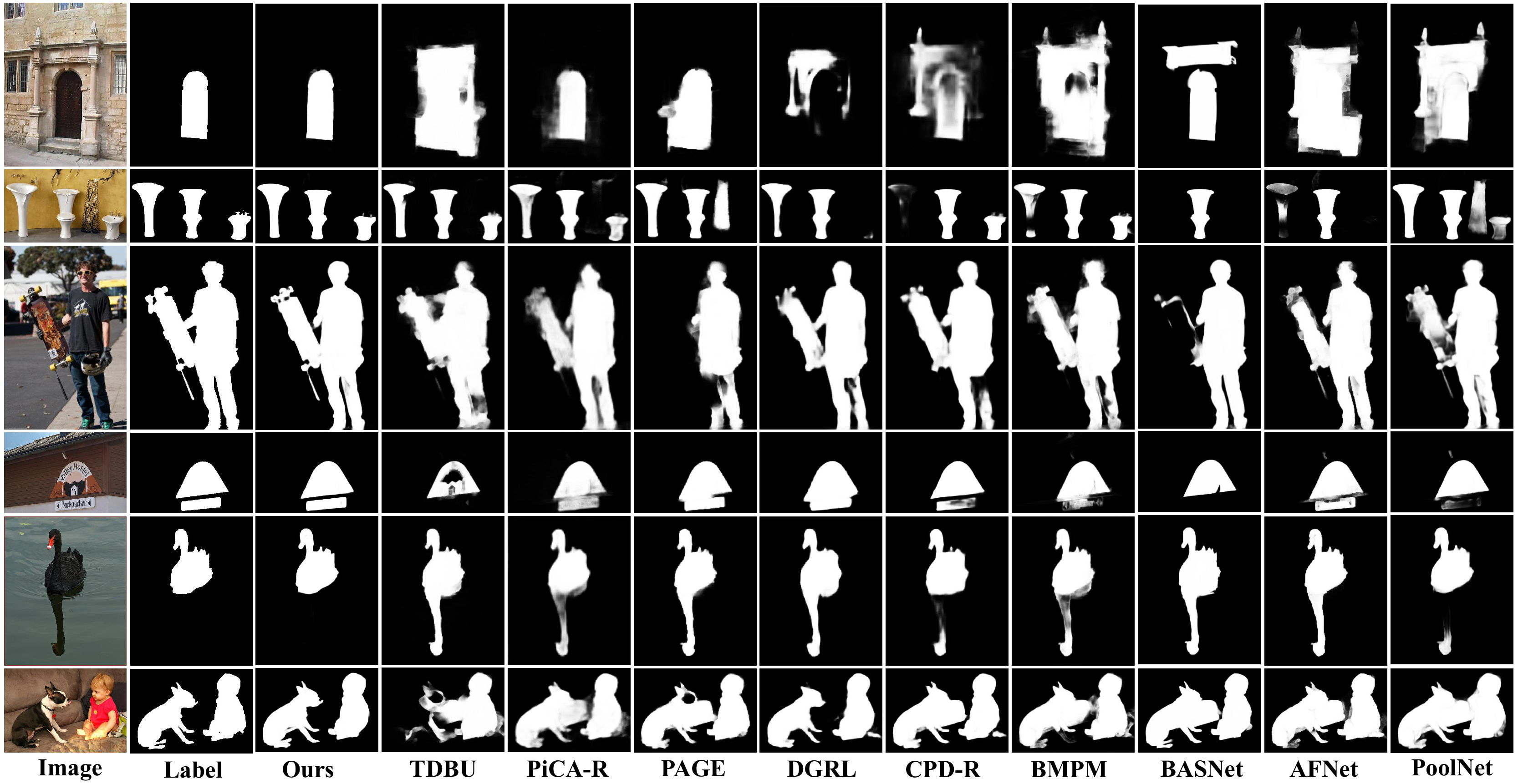}
  \caption{Visual comparison of different algorithms. Each row represents one image and corresponding saliency maps. Each column represents the predictions of one method. Apparently, our method is good at dealing with cluttered background and producing more accurate and clear saliency maps.}
  \label{Sample}
\end{figure*}

%%%%%%%%%%%%%%%%%%%%%%%%%%%%%%%%%%%%%%%%%%%%%%%%%%%%%%%%%%%%%%%%%%%%%%%%%%%%%%%%%%%%%%%%%%%%%%%%%%%%%%
\subsection{Datasets and Evaluation Metrics}
To evaluate the proposed method, six popular benchmark datasets are adopted, including ECSSD~\cite{ECSSD} with 1000 images, PASCAL-S~\cite{PASCALS} with 850 images, HKU-IS~\cite{HKUIS} with 4447 images, DUT-OMRON~\cite{DUTO} with 5168 images, DUTS~\cite{DUTS} with 15572 images and THUR15K~\cite{THUR15K} with 6232 images. Among them, \textbf{DUTS} is the largest saliency detection benchmark, which contains 10,553 training images (\textbf{DUTS-TR}) and 5,019 testing images (\textbf{DUTS-TE}). \textbf{DUTS-TR} is used to train the model, other datasets for evaluation. In addition, we also measure the model performance on the challenging SOC dataset~\cite{SOC} of different attributes. Five metrics are used to evaluate the performance of our model and existing state-of-the-art methods. The first metric is the mean absolute error (MAE), as shown in Eq.~\ref{mae}, which is widely adopted in~\cite{RAS,DSS,C2SNet,PICANet}. Mean $F$-measure ($mF$), $E$-measure ($E_\xi$)~\cite{Emeasure}, weighted $F$-measure ($F_\beta^\omega$) and $S$-measure ($S_\alpha$) are also widely used to evaluate saliency maps. In addition, precision-recall (PR) and $F$-measure curves are drawn to show the overall performance.
\begin{eqnarray}
  MAE = \frac{1}{H \times W} \sum_{i=1}^{H}\sum_{j=1}^{W}|P(i,j)-G(i,j)|
  \label{mae}
\end{eqnarray}
where $P$ is the predicted map and $G$ is the ground truth.

\subsection{Implementation Details} 
The proposed model is trained on DUTS-TR and tested on the above mentioned six datasets. For data augmentation, we use horizontal flip, random crop and multi-scale input images. ResNet-50, pretrained on ImageNet, is used to initialize the backbone ({\it i.e.}, block1 to block5) and other parameters are randomly initialized. We set the maximum learning rate to 0.005 for ResNet-50 backbone and 0.05 for other parts. Warm-up and linear decay strategies are used. The whole network is trained end-to-end by stochastic gradient descent (SGD). Momentum and weight decay are set to 0.9 and 0.0005, respectively. Batchsize is set to 32 and maximum epoch is set to 48. During testing, each image is simply resized to 352 x 352 and then fed into the network to get prediction without any post-processing. It is worth noting that the output saliency maps are used as the predictions rather than the addition of predicted body and detail maps.

\begin{table*}
  \caption{Performance on SOC~\cite{SOC} of different attributes. Each row represents one attribute and we report the mean $F$-measure scores of LDF and state-of-the-art methods. The last row shows the whole performance on the SOC dataset. The best and the second best results are highlighted in {\color{red}red} and {\color{blue}blue} respectively.}
  \label{Attribute}
  \renewcommand\tabcolsep{4.0pt}
  \renewcommand\arraystretch{1.1}
  \centering
  \begin{tabular}{cccccccccccccc}
    \hline 
    \hline
    Attr & PiCA-R & BMPM & R$^3$Net & DGRL & RAS & AFNet & BASNet & PoolNet & CPD-R & EGNet-R  & SCRN & Ours\\
    \hline
    \hline
     AC  & 0.721 & 0.727 & 0.659 & 0.744 & 0.664 & 0.763 & {\color{blue}0.773} & 0.746 & 0.765 & 0.739 & 0.770 & {\color{red}0.774}\\
     BO  & 0.706 & 0.802 & 0.637 & {\color{red}0.847} & 0.654 & {\color{blue}0.824} & 0.780 & 0.677 & 0.821 & 0.743 & 0.743 & 0.803\\
     CL  & 0.703 & 0.708 & 0.667 & 0.735 & 0.616 & 0.740 & 0.721 & 0.723 & 0.741 & 0.707 & {\color{blue}0.751} & {\color{red}0.772}\\
     HO  & 0.727 & 0.738 & 0.683 & 0.773 & 0.682 & {\color{blue}0.778} & 0.769 & 0.768 & 0.766 & 0.747 & 0.775 & {\color{red}0.807}\\
     MB  & 0.779 & 0.757 & 0.669 & 0.809 & 0.687 & 0.794 & 0.791 & 0.784 & 0.810 & 0.741 & {\color{blue}0.815} & {\color{red}0.840}\\
     OC  & 0.692 & 0.711 & 0.625 & 0.724 & 0.608 & 0.730 & 0.721 & 0.713 & {\color{blue}0.741} & 0.699 & 0.732 & {\color{red}0.756}\\
     OV  & 0.778 & 0.783 & 0.677 & 0.797 & 0.666 & {\color{blue}0.805} & 0.802 & 0.774 & 0.799 & 0.768 & 0.801 & {\color{red}0.820}\\
     SC  & 0.678 & 0.702 & 0.626 & 0.725 & 0.645 & 0.711 & 0.713 & 0.723 & 0.726 & 0.708 & {\color{blue}0.738} & {\color{red}0.774}\\
     SO  & 0.569 & 0.588 & 0.546 & 0.618 & 0.560 & 0.615 & 0.619 & 0.631 & 0.635 & 0.605 & {\color{blue}0.639} & {\color{red}0.676}\\
    \hline
     Avg & 0.662 & 0.673 & 0.611 & 0.698 & 0.608 & 0.700 & 0.697 & 0.694 & 0.709 & 0.680 & {\color{blue}0.710} & {\color{red}0.739}\\
    \hline
    \hline
  \end{tabular}
\end{table*}

\subsection{Ablation Studies}
%%%%%%%%%%%%%%%%%%%%%%%%%%%%%%%%%%%
\textbf{Number of Feature Interaction.} Tab.~\ref{iteration} shows the performance with different numbers of feature interaction. Compared with the baseline which has no feature interaction (Number=0), model with one feature interaction achieves better results. When the number is larger, the performance becomes worse. Because repeated feature interaction makes the network to grow too deeper and harder to optimize. So in all the following experiments, we set the number to 1 to balance the model optmization and performance.

\textbf{Different Combinations of Supervision.} Tab.~\ref{supervision} shows the performance with different combinations of supervision. From this table, combinations including detail label perform better than those including edge label, which demonstrates the effectiveness of detail label than edge label. In addition, combinations including body label perform better than those including saliency label (Sal). It confirms that without the interference of edges, center pixels can learn better feature representations.

\begin{figure*}[htb]
  \centering
  \subfigure[ ECSSD ]{\includegraphics[width=0.24\linewidth]{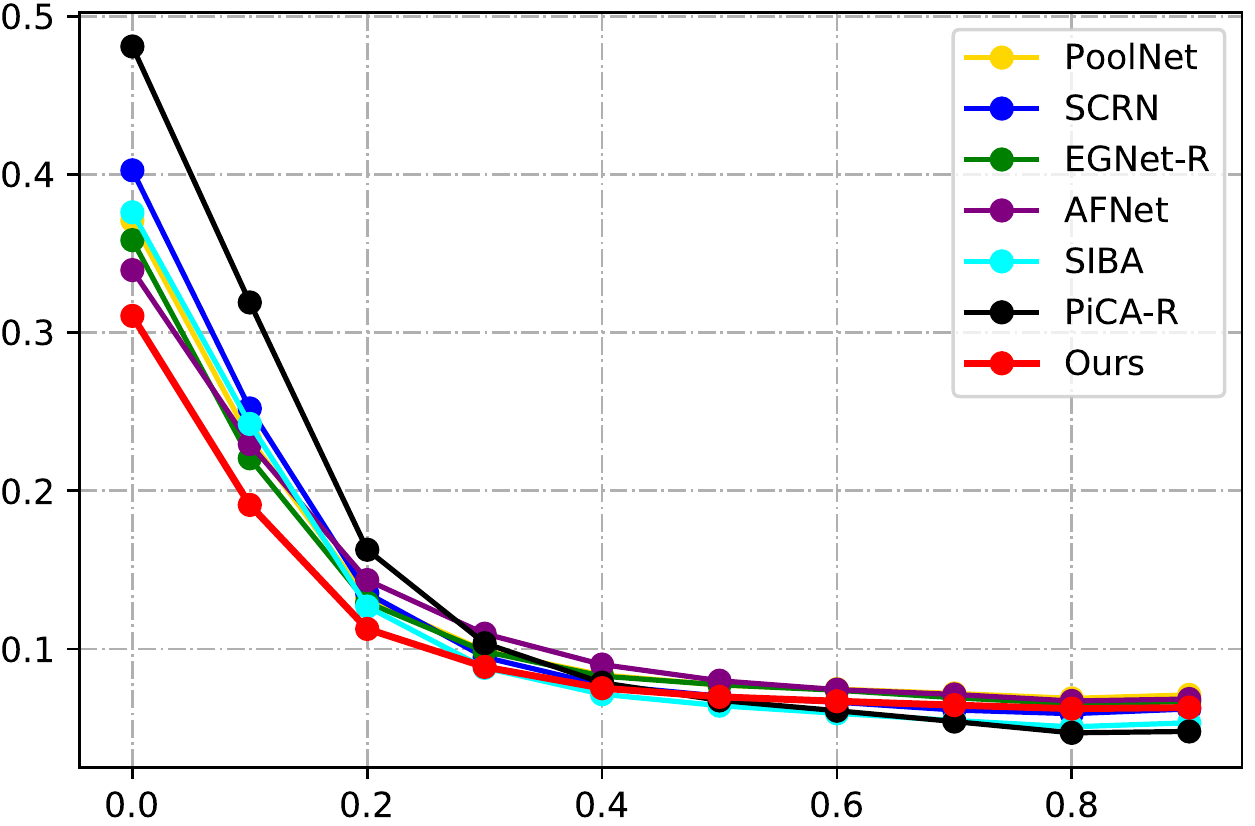}}
  \subfigure[  DUTS ]{\includegraphics[width=0.24\linewidth]{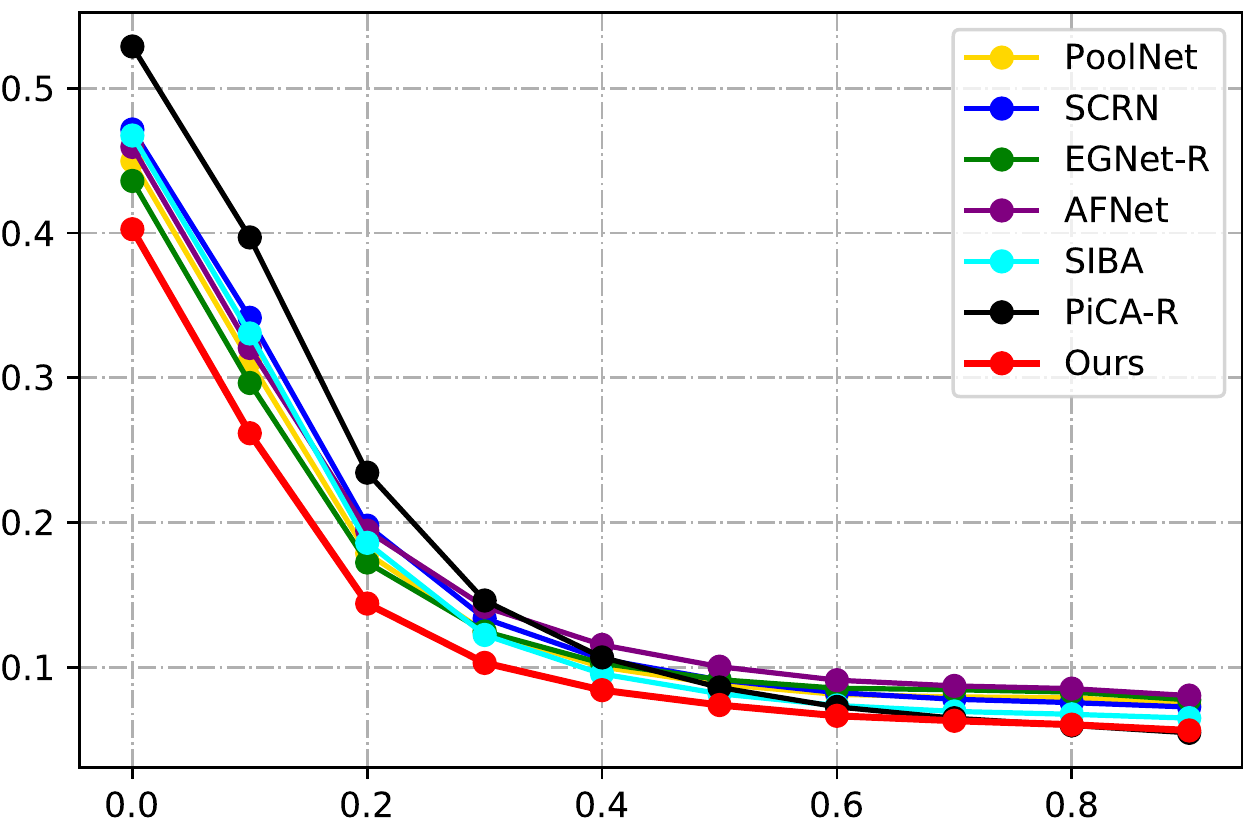}}
  \subfigure[HKU-IS ]{\includegraphics[width=0.24\linewidth]{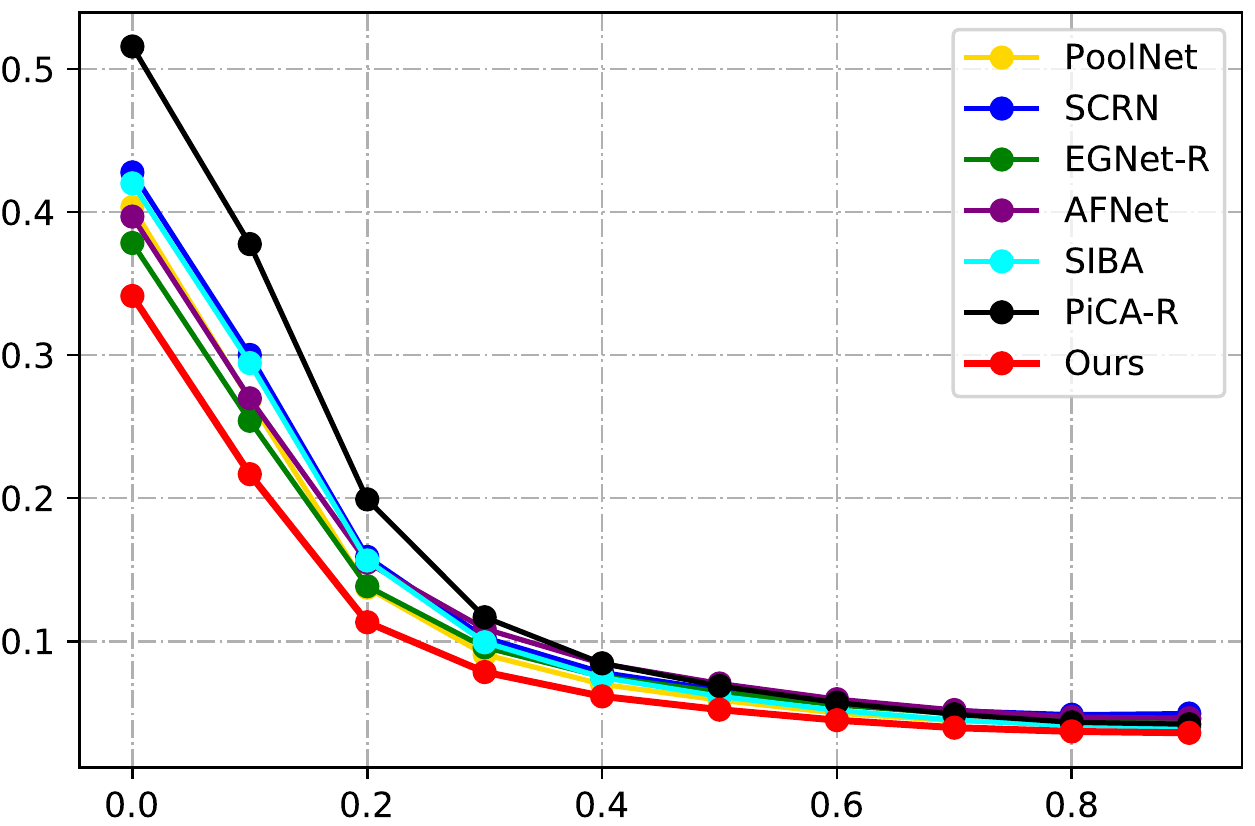}}
  \subfigure[THUR15K]{\includegraphics[width=0.24\linewidth]{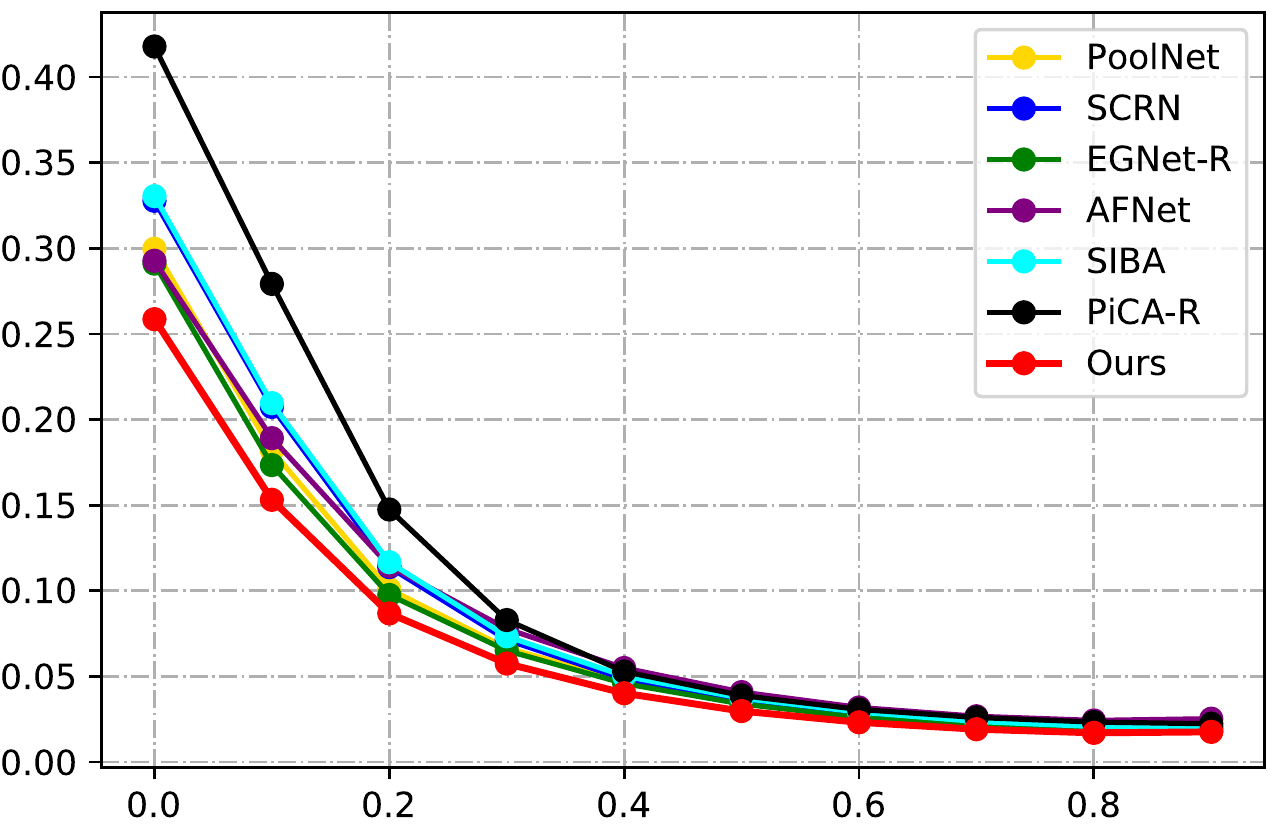}}
  \caption{Error-Distance distribution of different methods. The proposed method has the smallest error along the distance. Especially around edge areas, the proposed method performs much better.}
  \label{ErrorComparsion}
 \end{figure*}

\subsection{Comparison with State-of-the-arts}
\textbf{Quantitative Comparison.}
%%%%%%%%%%%%%%%%%%%%%%%%%%%%%%%%%%%
To demonstrate the effectiveness of the proposed method, 14 state-of-the-art SOD methods are introduced to compare, including BMPM~\cite{BMPM}, DGRL~\cite{DGRL}, R$^3$Net~\cite{R3Net}, RAS~\cite{RAS}, PiCA-R~\cite{PICANet}, AFNet~\cite{AFNet}, BASNet~\cite{BASNet}, CPD-R~\cite{CPD}, EGNet-R~\cite{EGNet}, PAGE~\cite{PAGE}, TDBU~\cite{TDBU}, SCRN~\cite{SCRN}, SIBA~\cite{SIBA} and PoolNet~\cite{PoolNet}. For fair comparison, we evaluate all the saliency maps provided by the authors with the same evaluation codes. We compare the proposed method with others in terms of MAE, $mF$ and $E_\xi$, which are shown in Tab.~\ref{Performance}. The best results are highlighted with red color. Obviously, compared with other counterparts, our method outperforms previous state-of-the-art methods by a large margin. Besides, Fig.~\ref{PRCurve} presents the precision-recall curves and $F$-measure curves on five datasets. As can be seen, the curves of the proposed method consistently lie above others. In addition, we calculate the Error-Distance distribution of different methods in Fig.~\ref{ErrorComparsion}, where predictions produced by the proposed method have the minimum error along distance, especially around the edge areas. 

\begin{table}
  \centering
  \caption{Performance with different numbers of feature interaction. Number=0 means two branches have no feature interaction.}
  \label{iteration}
  \renewcommand\tabcolsep{3.7pt}
  \renewcommand\arraystretch{1}
  \begin{tabular}{c|ccc|ccc}
    \hline
    \hline
    \multirow{2}{*}{Number} & \multicolumn{3}{c|}{THUR15K} & \multicolumn{3}{c}{DUTS-TE} \\
                       & MAE  & $mF$ & $E_\xi$ & MAE & $mF$ & $E_\xi$  \\
    \hline
    0 & 0.069 & 0.751 & 0.834 & 0.038 & 0.839 & 0.897 \\
    1 & 0.064 & 0.764 & 0.842 & 0.034 & 0.855 & 0.910 \\
    2 & 0.066 & 0.756 & 0.837 & 0.035 & 0.849 & 0.903 \\
    3 & 0.068 & 0.753 & 0.834 & 0.037 & 0.842 & 0.897 \\
    \hline
    \hline
  \end{tabular}
\end{table}

\begin{table}
  \centering
  \caption{Comparison on different combinations of supervision. Body, detail, saliency and edge maps are used, respectively.}
  \label{supervision}
  \renewcommand\tabcolsep{2.1pt}
  \renewcommand\arraystretch{1}
  \begin{tabular}{c|ccc|ccc}
    \hline
    \hline
    \multirow{2}{*}{Label} & \multicolumn{3}{c|}{THUR15K} & \multicolumn{3}{c}{DUTS-TE} \\
                            & MAE  & $mF$ & $E_\xi$ & MAE & $mF$ & $E_\xi$  \\
    \hline
    Body + Detail & 0.064 & 0.764 & 0.842 & 0.034 & 0.855 & 0.910 \\
    Body + Edge   & 0.066 & 0.758 & 0.836 & 0.036 & 0.850 & 0.904 \\
    Sal  + Detail & 0.066 & 0.756 & 0.835 & 0.037 & 0.848 & 0.901 \\
    Sal  + Edge   & 0.070 & 0.752 & 0.827 & 0.039 & 0.844 & 0.895 \\
    \hline
    \hline
  \end{tabular}
\end{table}

\textbf{Visual Comparison.}
%%%%%%%%%%%%%%%%%%%%%%%%%%%%%%%%%%%
Some prediction examples of the proposed method and other state-of-the-art approaches have been shown in Fig.~\ref{Sample}. We observe that the proposed method not only highlights the correct salient object regions clearly, but also well suppresses the background noises. It is robust in dealing with various challenging scenarios, including cluttered background, manufactured structure and low contrast foreground. Compared with other counterparts, the saliency maps produced by the proposed method are clearer and more accurate. 

\textbf{Performance on SOC of Different Attributes.}
SOC~\cite{SOC} is a challenging dataset with multiple attributes. Images with the same attribute have certain similarity and reflect the common challenge in real world. We utilize this dataset to test the robustness of model under different scenes. Specifically, we evaluate the mean $F$-measure score of our model as well as 11 state-of-the-art methods. Each model will get nine scores under nine attributes. In addition, an overall score is calculated to measure the whole performance under all scenes. Tab.~\ref{Attribute} shows the scores. We can see the proposed model achieves the best results among most of attributes except ``BO", which indicates the good generalization of the proposed method. It can be applied in different challenging scenes.

\section{Conclusion}
%%%%%%%%%%%%%%%%%%%%%%%%%%%%%%%%%%%%%%%%%%%%%%%%%%%%%%%%%%%%%%%%%%%%%%%%%%%%%%%%%%%%%%%%%%%%%%%%%%%%%%
In this paper, we propose the label decoupling framework for salient object detection. By empirically showing that edge prediction is a challenging task in saliency prediction, we propose to decouple the saliency label into body map and detail map. Detail map helps model learn better edge features and body map avoids the distraction from pixels near edges. Supervised by these two kinds of maps, the proposed method achieves better performance than direct supervison with saliency maps. Besides, feature interaction network is introduced to make full use of the complementarity between body and detail maps. Experiments on six datasets demonstrate that the proposed method outperforms state-of-the-art methods under different evaluation metrics.

\section{Acknowledgement}
This work was supported in part by the National Key R\&D Program of China under Grant 2018AAA0102003, in part by National Natural Science Foundation of China: 61672497, 61620106009, 61836002, 61931008 and U1636214, and in part by Key Research Program of Frontier Sciences, CAS: QYZDJ-SSW-SYS013. Authors would like to thank Kingsoft Cloud for their helpful disscussion and free GPU cloud computing resource support.

{
\small
\bibliographystyle{bibstyle}
\bibliography{bibrefer}
}

\end{document}